\documentclass[letterpaper]{article} 
\usepackage[draft]{aaai2026}  
\usepackage{times}  
\usepackage{helvet}  
\usepackage{courier}  
\usepackage[hyphens]{url}  
\usepackage{graphicx} 
\usepackage{natbib}  
\usepackage{caption} 
\usepackage{algorithm}
\usepackage{algorithmic}

\usepackage{newfloat}
\usepackage{listings}
\DeclareCaptionStyle{ruled}{labelfont=normalfont,labelsep=colon,strut=off} 
\floatstyle{ruled}
\newfloat{listing}{tb}{lst}{}
\floatname{listing}{Listing}
\usepackage{multirow}
\usepackage{booktabs}
\usepackage{amsmath}
\usepackage{makecell}
\usepackage{amssymb}
\usepackage{cleveref}
\usepackage{marvosym}

\newcommand{\bestmetric}[2]{${\textbf{#1}}^{\pm#2}$}
\newcommand{\secondmetric}[2]{${\underline{#1}}^{\pm#2}$}

\newcommand{\metric}[2]{${#1}^{\pm#2}$}

\title{
InterMoE: Individual-Specific 3D Human Interaction Generation via Dynamic Temporal-Selective MoE
}
\author{
    Lipeng Wang\textsuperscript{\rm 1},
    Hongxing Fan\textsuperscript{\rm 2},
    Haohua Chen\textsuperscript{\rm 1},
    Zehuan Huang\textsuperscript{\rm 1},
    Lu Sheng\textsuperscript{\rm 1}\thanks{Corresponding author.}
}
\affiliations{
    \textsuperscript{\rm 1}School of Software, Beihang University\\
    \textsuperscript{\rm 2}School of Computer Science and Engineering, Beihang University\\

    wanglipeng@buaa.edu.cn
}

\begin{document}

\maketitle

\begin{abstract}

Generating high-quality human interactions holds significant value for applications like virtual reality and robotics.
However, existing methods often fail to preserve unique individual characteristics or fully adhere to textual descriptions. 
To address these challenges, we introduce InterMoE, a novel framework built on a Dynamic Temporal-Selective Mixture of Experts. 
The core of InterMoE is a routing mechanism that synergistically uses both high-level text semantics and low-level motion context to dispatch temporal motion features to specialized experts. 
This allows experts to dynamically determine the selection capacity and focus on critical temporal features, thereby preserving specific individual characteristic identities while ensuring high semantic fidelity.
Extensive experiments show that InterMoE achieves state-of-the-art performance in individual-specific high-fidelity 3D human interaction generation, reducing FID scores by 9\% on the InterHuman dataset and 22\% on InterX.

\end{abstract}

\begin{links}
    \link{Code}{https://github.com/Lighten001/InterMoE}
\end{links}

\section{Introduction}

The generation of realistic and expressive Human Interaction has emerged as an important research area, propelled by rapid advancements in motion synthesis techniques. Faithfully modeling the complex joint motion between two individuals is crucial for a multitude of downstream applications, including but not limited to computer animation, virtual reality, and game development.

Recent works have made significant progress in generative human interaction. 
Some work~\cite{liang2024intergen,javedintermask} uses cross-attention to fuse features between interacting individuals. 
However, the subsequent uniform processing of these fused features by standard feed-forward networks (FFNs) tends to suppress individual characteristics, resulting in homogenized motions.
An alternative approach~\cite{li2024twoinone, wang2025TIMotion} concatenates individual features to jointly generate motion for both individuals; however, this method often suffers from identity confusion due to the absence of explicit identity constraints, which can cause characters to swap roles or positions during interaction erroneously.

\begin{figure*}[t]
\centering
\includegraphics[width=0.99\textwidth]{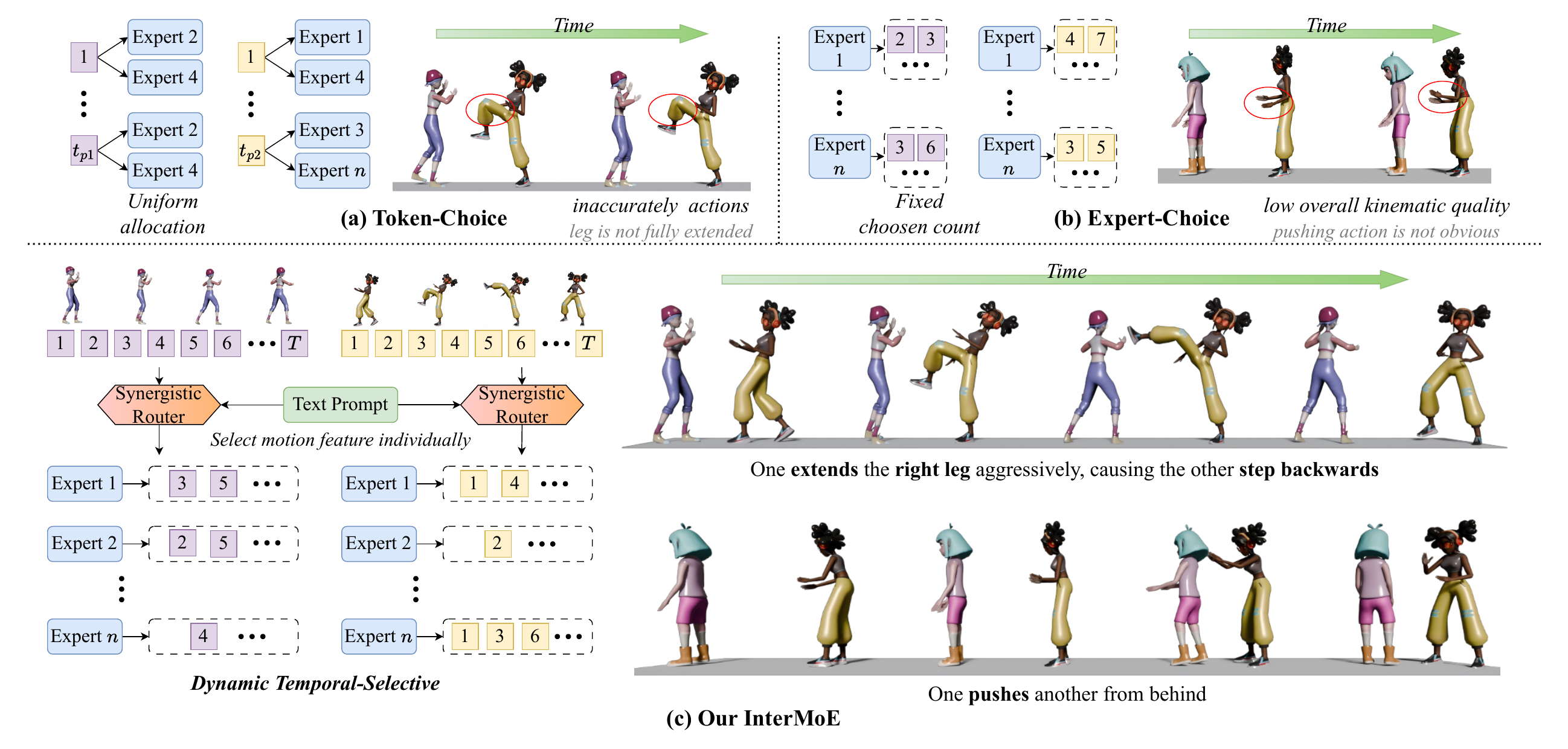}
\caption{
Compared with conventional MoE mechanisms, Token-Choice inaccurately generates the ``\textit{extends}'' action, and Expert-Choice has low overall kinematic quality. Our framework leverages the Synergistic Router and Dynamic Temporal Selection mechanism to generate 3D human interactions that exhibit both high semantic fidelity and robust preservation of individual-specific characteristics.
}
\label{fig:data_analysis}
\end{figure*}

To address the core challenge of preserving distinct identities, we argue that the inherent complexity of simultaneously modeling individual-specific characteristics and the joint motion between persons necessitates a modular approach handled by specialized sub-modules. We identify the Mixture of Experts (MoE) architecture~\cite{shazeer2017outrageously, lepikhin2020gshard} as a promising paradigm for this purpose. By design, MoE enables differentiated expert allocation by routing inputs based on their distinct motion characteristics, allowing for the development of specialists for each individual's unique motion patterns. This approach naturally mitigates identity confusion and homogenization.
Prevailing MoE routing strategies fall into two main categories. In the Token-Choice paradigm~\cite{fei2024scaling}, each token selects a fixed number of experts for processing. However, this uniform assignment fails to account for the varying importance across temporal features. The second category is Expert-Choice~\cite{sun2024ecditscalingdiffusiontransformers}, where experts select a fixed number of the most salient tokens. Yet, this fixed-capacity approach can limit expert utility, especially when modeling complex interactions.

In this work, we propose InterMoE, a novel framework that introduces a Dynamic Temporal-Selective MoE to generate high-fidelity, individual-specific 3D human interactions. Our framework is built upon two key innovations: a Synergistic Router that directs information based on both semantic and kinematic cues, and a Dynamic Temporal-Selection mechanism that empowers experts to focus on critical temporal features.
Specifically, the Synergistic Router leverages both high-level semantics from text and low-level kinematic features to guide routing decisions. This dual guidance ensures information is dispatched to the most appropriate experts, strengthening the alignment between the textual description and the generated motion. Building on this, the Dynamic Temporal-Selection mechanism enables each expert to dynamically determine the selection capacity and proactively identify the most salient temporal features, which effectively addresses the non-uniform temporal importance of the interactive motion sequence.

In summary, our contributions are as follows:

\begin{itemize}
    \item We propose InterMoE, a novel diffusion-based MoE framework for text-driven 3D human interaction generation, achieving notable improvements in individual-specific characteristics, semantic fidelity, and overall quality of interaction.

    \item We introduce the Dynamic Temporal-Selective MoE, a new paradigm tailored for generating 3D human interactions. It leverages a Synergistic Router that leverages both semantic and kinematic features for precise routing, and a Dynamic Temporal-Selection mechanism that empowers experts to dynamically focus on critical temporal features across varying noise levels.

    \item We conduct extensive experiments to demonstrate the effectiveness of our proposed framework. Furthermore, competitive results on the single-human motion generation task validate its strong generalization.
\end{itemize}
\section{Related Works}

\paragraph{Human Motion Generation}
Synthesizing single-person motion has gained interest, driven by large motion capture datasets and advancements in generative modeling techniques like Diffusions~\cite{tevet2023mdm, tseng2023edge, zhang2022motiondiffuse, chen2023mld, kong2023priority, lou2023diversemotion, zhang2024motiondiffuse} and Autoregressive models~\cite{guo2022tm2t, zhang2023motiongpt, jiang2023motiongpt, lucas2022posegpt, gong2023tm2d, pinyoanuntapong2024mmm}.
Recent works have explored various motion representations.
TM2T~\cite{guo2022tm2t} applies VQ-VAE~\cite{vqvae2018} to human motion data.
MoMask~\cite{guo2024momask} reduced quantization errors via residual quantization. While MotionStreamer~\cite{xiao2025motionstreamer} introduces a causal convolution to enforce temporal causality.
SALAD~\cite{hong2025salad} utilizes skeletal graph convolution to capture the spatial structure of human movement. Our work is motivated by these prior studies.

\paragraph{Human Interaction Generation}
Compared to single-human motion generation, human interaction generation is more challenging, as it requires accurately modeling interactions between individuals.
Recently, ComMDM~\cite{shafir2023human} trained a small neural network to bridge two single-person motion diffusion model copies on a limited interaction dataset.
InterGen~\cite{liang2024intergen} introduced a large-scale text-annotated two-person interaction dataset. And it proposed an interaction diffusion model that simultaneously denoises both individuals’ motions.
The in2IN~\cite{ruiz2024in2in}  further advances the field
by introducing a diffusion model that conditions motion generation not only on overall interaction descriptions but also on individual actions.
InterMask~\cite{javedintermask} employs a spatial-temporal transformer to autoregressively generate interactions. 
TIMotion~\cite{wang2025TIMotion} models temporal and interactive dynamics in interactions. Although the above-mentioned methods have achieved promising results, they still show limitations in differentiating the identity-specific characteristics of each individual and remain semantically faithful.

\begin{figure*}[t]
\centering
\includegraphics[width=0.99\textwidth]{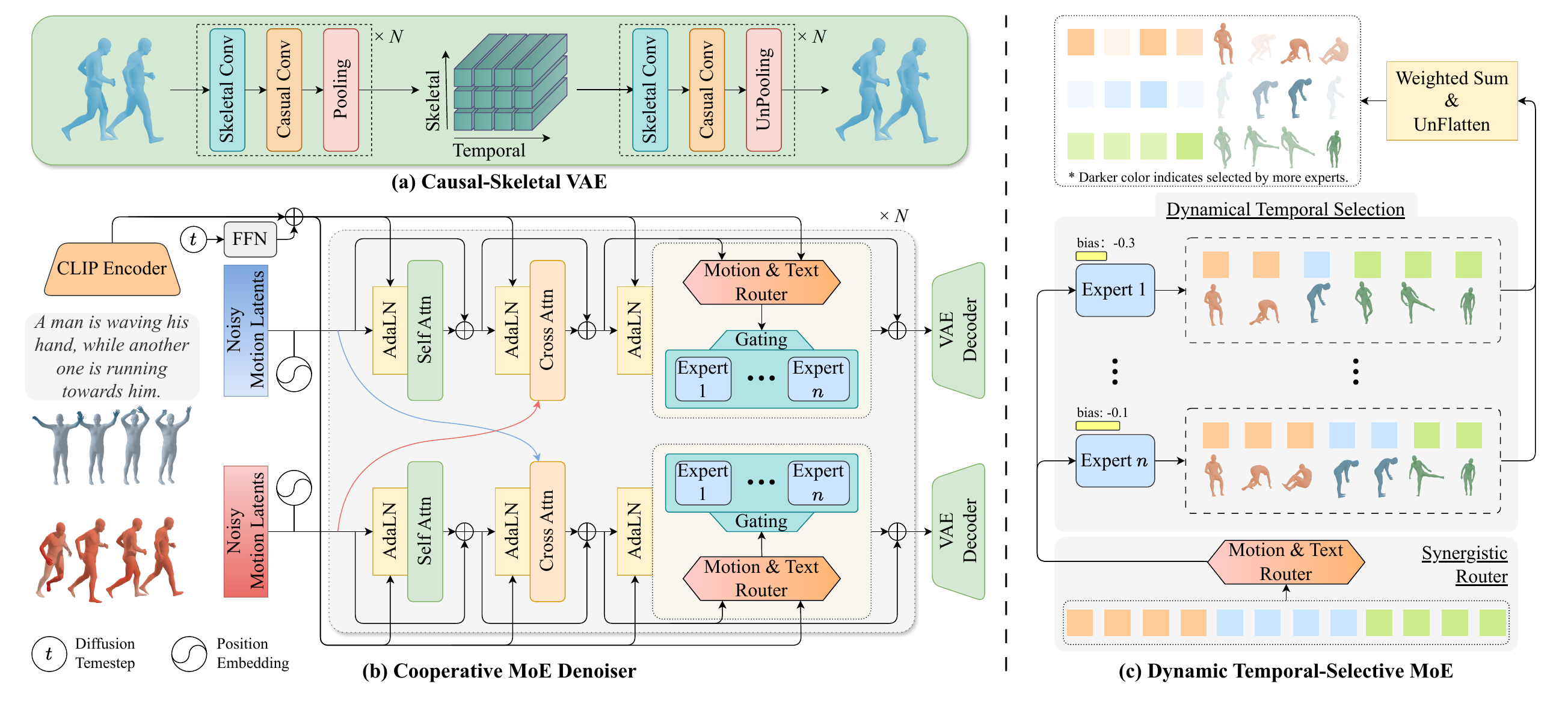}
\caption{
The overall framework of the InterMoE. 
(a) Causal-Skeletal VAE to encode/decode individual motions;
(b) Two Cooperative MoE Denoisers to interactively perform denoising;
(c) Our proposed Synergistic Router and  Dynamic Temporal-Selective Expert mechanism. The router guides multiple experts to select and process critical temporal features of the motion sequence dynamically.
}
\label{fig:method_main}
\end{figure*}

\paragraph{Mixture-of-Experts}
Mixture-of-Experts~\cite{shazeer2017outrageously, lepikhin2020gshard} paradigm has become a powerful and efficient strategy for scaling models while maintaining manageable computational costs by selectively activating expert subsets.
This approach has made remarkable success in Large Language Models and multimodal large models, such as ~\cite{deepseekai2024deepseekv3technicalreport, minimax2025minimax01scalingfoundationmodels, chen2023octavius, muennighoff2024olmoeopenmixtureofexpertslanguage}.
Recent works have explored incorporating MoE architectures into diffusion models. MEME~\cite{lee2023multiarchitecturemultiexpertdiffusionmodels}, eDiff-I~\cite{balaji2022eDiff-I}, and ERNIE-ViLG 2.0~\cite{feng2023ernievilg20improvingtexttoimage} restrict expert selection to a specific timestep. SegMoE~\cite{gupta2024segmoe} and DiT-MoE~\cite{fei2024scaling} suffer from significant expert utilization imbalance due to isolated token processing. While EC-DiT~\cite{sehwag2024stretchingdollardiffusiontraining, sun2024ecditscalingdiffusiontransformers} dispatch consequential tokens to each expert.
Su~\cite{su2025moe} proposes a loss-free dynamical routing method.
DiffMoE~\cite{shi2025diffmoe} utilizes a batch-level global token pool and dynamically adapting computation. Our work is informed by the foundation laid by these prior studies.

\section{Methods}

We target the task of text-driven synthesis of 3D human interaction. Given a textual description, our model generates a set of motion sequences $\textbf{m}_i, i\in \{1,2\}$, where $\textbf{m}_i$ represents the 3D motion for an individual $i$. Each 3D motion $\textbf{m}_i \in \mathbb{R}^{T \times J \times d}$ is composed of $T$ frames, with each frame describing a pose via $J$ joints, each represented by a $d$-dimensional feature vector. 
The overall pipeline of our framework is illustrated in Figure~\ref{fig:method_main}.
In this section, we first briefly introduce the Causal-Skeletal VAE and the Cooperative MoE Denoiser used in the diffusion model in section~\ref{subsec:inter-ldm}. Then we give a detailed explanation of the core MoE design in section~\ref{subsec:intermoe}.

\subsection{Interaction Latent Diffusion}\label{subsec:inter-ldm}
\paragraph{Causal-Skeletal VAE}

Recent works have demonstrated the efficacy of applying the graph convolution to the human joint topology for extracting skeletal features~\cite{hong2025salad}, while the causal convolutions preserve temporal causality and enable efficient encoding of sequential data~\cite{yu2023language}. Building upon these findings, we devise a hierarchical encoder-decoder architecture to embed the single-person motion. As shown in Figure~\ref{fig:method_main}(a), we first use skeletal convolutions to capture the complex intra-frame human joint dependencies. The resulting then fed into a causal convolution to model the inter-frame temporal dynamics and causal relationships. Finally, a pooling layer compresses these skeletal-temporal features into a compact representation. This design yields a lightweight yet highly efficient representation for our motion VAE.

Specifically, when feeding motion embeddings into the denoising network, we flatten the joint dimensions the same as InterGen~\cite{liang2024intergen} (i.e., reshape $\textbf{m}_i^\text{ori} \in \mathbb{R}^{T \times J\times d}$ to $\textbf{m}_i \in \mathbb{R}^{T \times D_m}$, where $D_m = J \times d$).

\paragraph{Cooperative MoE Denoiser} 

The architecture of our denoising network inherits the design of InterGen~\cite{liang2024intergen}. It utilizes two share-weight cooperative Denoisers to process the human interaction. Each denoiser is composed of a series of transformer blocks. As shown in Figure~\ref{fig:method_main}(b), each block contains three core components: 
(1) a Self-Attention Layer to model intra-individual temporal relationships; 
(2) a Cross-Attention Layer that conditions on the motion features of the interaction partner to model inter-individual dynamics; 
(3) and our proposed MoE Block. Furthermore, we integrate the denoising timestep and text-conditional information into the network via Adaptive Layer Normalization before all attention layers and the MoE Block.

\subsection{InterMoE}\label{subsec:intermoe}

Prior work~\cite{wang2025TIMotion} has shown that relying solely on cross-attention mechanisms is often insufficient for preserving distinct identities in human interaction synthesis. To overcome this limitation and to enhance the overall quality and semantic fidelity of generated sequences. 
Here, we introduce the core of InterMoE, which consists of two key components: the Synergistic Router and the Dynamic Temporal Selection. These components enable the generation of high-fidelity identity-preserving 3D human interaction.

\subsubsection{Synergistic Router}

As shown in Figure~\ref{fig:method_main}(b) and ~\ref{fig:method_main}(c), our synergistic router operates on the motion features $\textbf{m}_i$ and the text condition $c_t$. 
We employ a motion router to generate routing logits based on the unique kinematic signatures of each individual. Concurrently, a parallel text router takes the text features as input and computes the semantic routing logits. 

Furthermore, we identify that 
an instance-centric routing scope prevents routers from perceiving the heterogeneity of noise levels across different samples in a batch and hinders the discovery of global motion patterns. To overcome this, we introduce a batch-level routing strategy. Specifically, we flatten the input motion feature along its batch dimension (i.e., reshape $\textbf{m}_i \in \mathbb{R}^{B\times T \times D_m}$ to $\textbf{m}_i^\text{flat} \in \mathbb{R}^{S \times D_m}$, where $S = B \times T$) to create a batch-level temporal feature pool.
Then calculate the routing logits $\mathbf{R}_e$ for each expert $e$.
Note that $\textbf{m}_i^\text{flat} = [m_{s,i}^\text{flat}], \ s \in \{1, \dots , S\}$, we have
\begin{align}
\mathbf{R}_{e,s,i}^\text{motion} &=  \textbf{Router}_{e}^\text{motion}({m}_{s,i}^\text{flat}), \  {m}_{s,i}^\text{flat} \in \mathbb{R}^{1\times D_m} \label{eq:motion-router}\\
\mathbf{R}_{e}^\text{text} &= \textbf{Router}_{e}^\text{text}(c_t), \  c_t \in \mathbb{R}^{1 \times D_t}
\end{align}
These two sets of logits are subsequently fused via a weighted summation to produce the final logits $\mathbf{R}_{e,s,i}^\text{comb}$. 
\begin{equation}
\mathbf{R}_{e,s,i}^\text{comb} = \alpha \mathbf{R}_{e,s,i}^\text{motion} + (1-\alpha) \mathbf{R}_{e}^\text{text}
\end{equation}
In experiments, we set $\alpha=0.5$. Through this design, routers are guided by both individual-specific dynamics and high-level semantic intent.
By utilizing a batch-level temporal feature pool, the router can perform a more nuanced analysis of motion features and fully leverage the information inherent to different noise levels during allocation.

\subsubsection{Dynamic Temporal Selection}

The conventional Token-Choice paradigm treats all tokens uniformly, which overlooks the non-uniform salience of temporal features in interactive motion sequences. To address this, we propose a Dynamic Temporal Selection mechanism, as shown in Figure~\ref{fig:method_main}(c). 
This mechanism empowers each expert to proactively select critical tokens from the entire batch-level temporal feature pool for processing. 
Furthermore, we remove constraints on each expert to a fixed capacity of selecting top-K features. Instead, we introduce a dynamic selection mechanism. 
Specifically, within each MoE module of our network, we associate every expert with a learnable bias parameter $b_e$.
Notably, each Cooperative MoE Denoiser is dedicated to processing a single individual within the interaction, so omitting the annotation $i$ for different individuals, we have:
\begin{align}
\mathbf{M}_{e,s} & =\texttt{sigmoid}(\mathbf{R}^\text{comb}_{e,s}) + {b}_e \\
\mathbf{A}_{e,s} & =\texttt{softmax}(\mathbf{R}^\text{comb}_{e,s}) \\
\mathbf{G}_{e,s} & = 
\begin{cases}
\mathbf{A}_{e,s}, & \mathbf{M}_{e,s} > 0,\\
0, & \text{otherwise}.
\end{cases} 
\end{align}
where $\mathbf{G}_{e,s}$ is the final gating weight, $\mathbf{M}_{e,s}$ determines to select this feature or not for experts. Since the $\texttt{sigmoid}$ function maps values to $(0, 1)$, this bias ${b}_e$ is constrained in $(-1, 0)$, which determines the number of features that the expert $e$ will process, where a larger bias (i.e., a value closer to 0) corresponds to a higher capacity. 
Critically, these bias parameters ${b}_{e}$ are optimized during training.
With $K_e^\text{exp}$ as a hyperparameter, in a training step, we count the number of selected features $ K_{e}^\text{select}$ for each expert:
\begin{align}
K_{e}^\text{select} &= \textbf{Count}(\mathbf{M}_{e,s} > 0), \ s \in \{1,...,S\}.
\end{align}
After this training step, we update bias ${b}_{e}$ as:
\begin{align}
\Delta b_e &=  \text{sign}(K_{e}^\text{select} - K_{e}^\text{exp}) \\
{b}_{e} &\leftarrow b_e - \sigma ~ \Delta b_{e}
\end{align}
In experiments, we set $\sigma=1 \times e^{-4}$, As training converges, the $\Delta b_e$ term is driven toward zero, ensuring that $\mathbf{E}(K_e^\text{select})$ approximates to $K_e^\text{exp}$, which facilitates a dynamic but stable allocation. Finally, we obtain the output feature $ {m}_s^\text{out} $ from the MoE block:
\begin{align}
{m}_s^\text{out} & = \sum\limits_{e=1}^{N} \mathbf{G}_{e,s} E({m}_s), \  {m}_s \in \mathbb{R}^{1\times D}, \  s \in \{1,\dots,S\}.
\end{align}
This design enables each expert to dynamically determine the feature selection capacity and select their preferred temporal motions across all sequences within the entire batch. This global perspective yields a dual advantage: First, it endows the experts with noise-level awareness, enabling more robust feature selection at different denoising stages during inference. Second, it significantly enhances the experts' ability to identify universally critical temporal features through exposure to a more diverse set of motion samples.

\begin{figure*}[ht]
\centering
\includegraphics[width=0.99\textwidth]{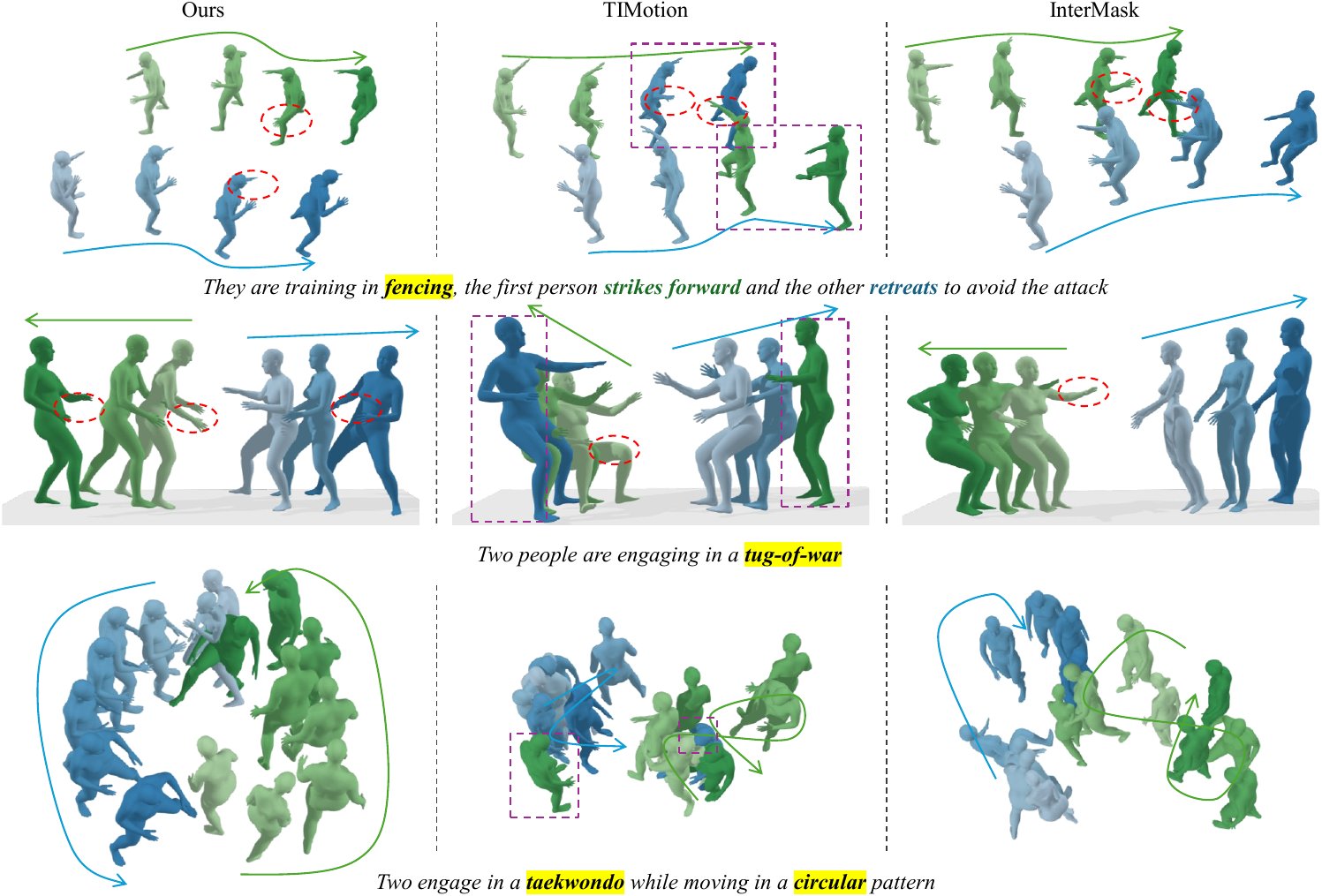}
\caption{
Qualitative comparisons with TIMotion~\shortcite{wang2025TIMotion} and InterMask~\shortcite{javedintermask}.
Arrowed lines mark the trajectories of motion, Red circles indicate key actions that align with the text, and Purple boxes highlight the identity confusion error.
}
\label{fig:qual_cmp}
\end{figure*}

\section{Experiments}\label{sec:exp}
\subsection{Experimental setup}

\paragraph{Datasets}
We adopt two datasets to evaluate our method for the text-conditioned human interaction generation task: InterHuman~\cite{liang2024intergen} and InterX~\cite{xu2024interx}. 
The InterHuman dataset contains 7,779 interaction sequences, and InterX contains 11,388, each paired with 3 distinct textual annotations. 
InterHuman follows the AMASS~\cite{mahmood2019amass} skeleton representation with 22 joints, including the root joint. Each joint is represented by $ \{pos, vec, rot\} $, where $pos \in \mathbb{R}^3$ is the global position, $vec \in \mathbb{R}^3$ is the global velocity, and $rot \in \mathbb{R}^6$ is the local 6D rotation of each joint, rendering $\mathbf{m}_p \in \mathbb{R}^{N \times 22 \times 12}$. 
InterX follows the SMPL-X~\cite{SMPL-X:2019} skeleton representation, comprising 55 body, hands, and face joints. Each joint is represented by $\{pos, vec, rot\}$  where $pos \in \mathbb{R}^3$ is the global position, $vec \in \mathbb{R}^3$ is the global velocity, and $rot \in \mathbb{R}^6$ is the local 6D rotation of each joint, rendering $\mathbf{m}_p \in \mathbb{R}^{N \times 55 \times 12}$.

\paragraph{Metrics}
We employ the same evaluation metrics as T2M~\cite{guo2022generating} and InterGen~\cite{liang2024intergen}, which are as follows: 
{
(1) Frechet Inception Distance (FID). 
(2) R-Precision. 
(3) Diversity. 
(4) Multimodality (MModality). 
(5) Multi-modal distance (MM Dist). 
}
Detailed explanations of the metrics can be found in the Appendix.

\paragraph{Implementation Details}
Our framework was trained on two NVIDIA RTX3090 GPUs. We used the AdamW optimizer with betas of $(0.9, 0.999)$, a weight decay of $2 \times 10^{- 5}$, a maximum learning rate of $ 1 \times 10^{- 4} $, and a cosine LR schedule with 10 linear warm-up epochs. For InterHuman dataset, the VAE was trained for 100 epochs with batch size of 256, and the denoiser was trained for 1000 epochs with batch size of 64, respectively. For Inter-X dataset, the VAE was trained for 500 epochs with batch size of 256, and the denoiser was trained for 2000 epochs with batch size of 64, respectively. We trained the denoiser with 1000 diffusion steps, employing 50 steps for DDIM sampling during inference. For the CFG weight, we set $w = 3.5$ unless mentioned otherwise.

\subsection{Comparative experiments}
\begin{table*}[ht]\small
    \centering
    \setlength{\tabcolsep}{0.7mm}{
        \begin{tabular}{c   l c c c c c c c}
            \toprule
            \multirow{2}{*}{Datasets} & \multirow{2}{*}{Methods} & \multicolumn{3}{c}{R-Precision $\uparrow$} & \multirow{2}{*}{FID $\downarrow$} & \multirow{2}{*}{MM-Dist $\downarrow$} & \multirow{2}{*}{Diversity $\rightarrow$} & \multirow{2}{*}{\makecell[c]{Multi\\Modality} $\uparrow$} \\
            & & Top-1 & Top-2 & Top-3 &  &  &  &  \\

            \midrule

            \multirow{10}{*}{\textit{\makecell[c]{Inter\\Human}}}

            & Real motion & \metric{0.452}{.008} & \metric{0.610}{.009} & \metric{0.701}{.008} & \metric{0.273}{.007} & \metric{3.755}{.008} & \metric{7.948}{.064} & - \\
            
            \cmidrule{2-9}
            
            & T2M~\shortcite{guo2022generating} & \metric{0.238}{.012} & \metric{0.325}{.010} & \metric{0.464}{.014} & \metric{13.769}{.072} & \metric{5.731}{.013} & \metric{7.046}{.022} & \metric{1.387}{.076}  \\

            & MDM~\shortcite{tevet2023mdm} & \metric{0.153}{.012} & \metric{0.260}{.009} & \metric{0.339}{.012} & \metric{9.167}{.056} & \metric{7.125}{.018} & \metric{7.602}{.045} & \metric{\textbf{2.350}}{.080} \\   
        
            & ComMDM~\shortcite{shafir2023human} & \metric{0.223}{.009} & \metric{0.334}{.008} & \metric{0.466}{.010} & \metric{7.069}{.054} & \metric{6.212}{.021} & \metric{7.244}{.038} & \metric{1.822}{.052} \\

            & InterGen~\shortcite{liang2024intergen} & \metric{0.371}{.010} & \metric{0.515}{.012} & \metric{0.624}{.010} & \metric{5.918}{.079} & \metric{5.108}{.014} & \metric{7.387}{.029} & \secondmetric{2.141}{.063} \\
        
            & MoMat-MoGen~\shortcite{cai2024digital} & \metric{0.449}{.004} & \metric{0.591}{.003} & \metric{0.666}{.004} & \metric{5.674}{.085} & \metric{3.790}{.001}  & \metric{8.021}{.35} & \metric{1.295}{.023} \\

            & in2IN~\shortcite{ruiz2024in2in} & \metric{0.425}{.008} & \metric{0.576}{.008} & \metric{0.662}{.009} & \metric{5.535}{.120} & \metric{3.803}{.002}  & \secondmetric{7.953}{.047} & \metric{1.215}{.023} \\

            & InterMask~\shortcite{javedintermask} & \metric{0.449}{.004} & \metric{0.599}{.005} & \metric{0.683}{.004} & \secondmetric{5.154}{.061} & \metric{3.790}{.002}  & \bestmetric{7.944}{.033} & \metric{1.737}{.020}  \\

            & TIMotion~\shortcite{wang2025TIMotion} & \secondmetric{0.491}{.005} & \secondmetric{0.648}{.004} & \secondmetric{0.724}{.004} & \metric{5.433}{.080} & \secondmetric{3.775}{.001}  & \metric{8.032}{.030} & \metric{0.952}{.032}  \\

            \cmidrule{2-9}
            
            & \textbf{Ours} & \bestmetric{0.512}{.004} & \bestmetric{0.671}{.004} & \bestmetric{0.746}{.004} & \bestmetric{4.677}{.069} & \bestmetric{3.762}{.001}  & \metric{7.990}{.029} & \metric{0.964}{.028} \\
            
    \toprule
        \multirow{8}{*}{\textit{InterX}}
        
            & Real motion & \metric{0.429}{.004} & \metric{0.626}{.003} & \metric{0.736}{.003} & \metric{0.002}{.0002} & \metric{3.536}{.013} & \metric{9.734}{.078} & -  \\

            \cmidrule{2-9}
    
            & T2M~\shortcite{guo2022generating} & \metric{0.184}{.010} & \metric{0.298}{.006} & \metric{0.396}{.005} & \metric{5.481}{.382} & \metric{9.576}{.006} & \metric{2.771}{.151}  & \metric{2.761}{.042}\\   
            
            & MDM~\shortcite{tevet2023mdm} & \metric{0.203}{.009} & \metric{0.329}{.007} & \metric{0.426}{.005} & \metric{23.701}{.057} & \metric{{9.548}}{.014} & \metric{5.856}{.077} & \secondmetric{{3.490}}{.061} \\
            
            & ComMDM~\shortcite{shafir2023human} & \metric{0.090}{.002} & \metric{0.165}{.004} & \metric{0.236}{.004} & \metric{29.266}{.067} & \metric{6.870}{.017}  & \metric{4.734}{.067} & \metric{0.771}{.053} \\
    
            & InterGen~\shortcite{liang2024intergen} & \metric{{0.207}}{.004} & \metric{{0.335}}{.005} & \metric{{0.429}}{.005} & \metric{{5.207}}{.216} & \metric{9.580}{.011}  & \metric{{7.788}}{.208} & \metric{\textbf{3.686}}{.052} \\
    
            & InterMask~\shortcite{javedintermask} & \metric{0.403}{.005} & \metric{0.595}{.004} & \metric{0.705}{.005} & \metric{0.399}{.013} & \secondmetric{3.705}{.017} & \metric{9.046}{.073}  & \metric{2.261}{.081} \\
    
            & TIMotion~\shortcite{wang2025TIMotion} & \secondmetric{0.412}{.004} & \secondmetric{0.601}{.004} & \secondmetric{0.714}{.003} & \secondmetric{0.385}{.022} & \metric{3.706}{.015}  & \secondmetric{9.191}{.092} & \metric{2.437}{.069}  \\

            \cmidrule{2-9}
            
            & \textbf{Ours} & \bestmetric{0.427}{.003} & \bestmetric{0.612}{.004} & \bestmetric{0.721}{.004} & \bestmetric{0.297}{.015} & \bestmetric{3.605}{.016}  & \metric{9.109}{.083} & \metric{2.446}{.069}  \\

        \bottomrule
        \end{tabular}
    }

    \caption{
    Quantitative evaluation results on the test sets of InterHuman and Inter-X datasets. 
    $\uparrow$ and $\downarrow$ denote that higher and lower values are better, respectively, while $\rightarrow$ denotes that the values closer to the real motion are better. We run the evaluations 20 times. $\pm$ indicates a 95\% confidence interval.
    }
    \label{tab:cmp_main}
\end{table*}

\subsubsection{Quantitative Results}
Table~\ref{tab:cmp_main} shows quantitative comparisons of our InterMoE with previous methods on both InterHuman and InterX datasets. Following established practices~\cite{liang2024intergen, zhang2023t2m}, each experiment is conducted 20 times, and the reported metric values represent the mean with a 95\% statistical confidence interval. 
Our framework achieves state-of-the-art results on both InterHuman and InterX datasets. It records the lowest FID scores (4.677 on InterHuman and 0.297 on InterX), indicating superior realism and quality of generated interactions, and leads in R-Precision and MM-Dist, showing excellent semantic-fidelity. While our MultiModality is slightly lower than some methods, the high R-Precision and low MM-Dist emphasize that InterMoE prioritizes adherence to text over extreme diversity.

\subsubsection{Qualitative Comparisons}

In Figure~\ref{fig:qual_cmp}, we provide a qualitative comparison of interaction sequences generated by our InterMoE and prior state-of-the-art methods for the same text descriptions. 
Given the prompt ``\textit{They are training in fencing, the first person strikes forward and the other retreats to avoid the attack}'', existing methods like TIMotion exhibit significant identity confusion. Furthermore, both TIMotion and InterMask fail to render distinct offensive and defensive hand gestures or the precise forward and backward movements.
In the ``\textit{Two people are engaging in a tug-of-war}'' scenario, our InterMoE accurately synthesizes the hand-gripping-rope posture and the backward-leaning motion, whereas competitors fail to produce the correct kinematically plausible action. 
For a complex, 10-second interaction, ``\textit{Two engage in a taekwondo while moving in a circular pattern}'', InterMoE not only generates coherent sparring motions but also strictly adheres to the specified circular movement pattern. In contrast, TIMotion and InterMask disregard this moving constraint, producing only stationary sparring.
Collectively, these examples demonstrate that InterMoE generates higher-quality interactions with more precise semantic alignment and clearer individual-specific characteristics compared to prior methods.

\subsection{Ablation Studies and Analysis}

\subsubsection{Main Ablation}
\begin{table}[ht]\small
    \centering
    \setlength{\tabcolsep}{0.1mm}{
        \begin{tabular}{l c c c c c}
            \toprule
            \multirow{1}{*}{Methods} & \makecell[c]{R-Precision\\Top-1} & \multirow{1}{*}{FID $\downarrow$} & \multirow{1}{*}{MM-Dist $\downarrow$} \\

            \midrule

            \makecell[l]{Baseline\\(InterGen w/ CS-VAE)}& \metric{0.489}{.006} & \metric{5.251}{.086} & \metric{3.771}{.001} \\

            \midrule
            
            \multicolumn{4}{c}{
            \underline{
            \textit{Synergistic Router}
            }
            } \\
            
            w/o Motion \& Text Router & \metric{0.503}{.003} & \metric{4.782}{.066} & \metric{3.766}{.001} \\

            w/o Batch-level Routing & \metric{0.492}{.006} & \metric{6.036}{.072} & \metric{3.774}{.001} \\

            \midrule
            \multicolumn{4}{c}{
            \underline{
            \textit{Dynamic Temporal Selection}
            }
            } \\
            
            w/o Dynamic Selection & \metric{0.498}{.005} & \metric{6.242}{.070} & \metric{3.772}{.001} \\

            w/o Temporal-Selective & \metric{0.505}{.006} & \metric{5.195}{.070} & \metric{3.767}{.001} \\

            \midrule
            Ours Full  & \bestmetric{0.512}{.004}  & \bestmetric{4.677}{.069} & \bestmetric{3.762}{.001} \\

            \bottomrule
        \end{tabular}
    }

    \caption{
    Ablation study results on the InterHuman test set to verify key components of our InterMoE.
    }
    \label{tab:abl_components}
\end{table}

We begin by analyzing the impact of each key component.
For a fair comparison, we define the baseline as the InterGen framework integrated with the Causal-Skeletal VAE (CS-VAE).
As shown in Table~\ref{tab:abl_components}, we first validate the contributions of our Synergistic Router design. Removing the parallel motion and text Router leads to a decline in both R-Precision and FID, demonstrating that our dual router design is crucial for enhancing semantic fidelity and generation quality. More notably, when batch-level routing is disabled and routing decisions are confined to the instance level, the FID score degrades significantly. This strongly substantiates that enabling the router to perceive and leverage global batch information is indispensable for learning a high-quality generative distribution.
Next, we evaluate our temporal-selective expert mechanism. Disabling the dynamic selection and reverting to a fixed top-K features per expert also results in a substantial drop in FID, highlighting the superiority of learnable expert capacities. When removing the temporal-selective mechanism by uniformly assigning experts, its FID score is substantially worse. This performance gap highlights the critical importance of empowering experts to proactively select temporal features. 
Our full framework achieves the best performance across all metrics. These results indicate that our proposed Synergistic Router and Dynamic Temporal Selection mechanisms are not only effective individually but also work cooperatively to elevate the quality and fidelity of the generated interactions.

\begin{table}[t]\small
    \centering
    \setlength{\tabcolsep}{0.9mm}{
        \begin{tabular}{l c c c}
            \toprule
            {\makecell[l]{MoE Type}} & \makecell[c]{R-Precision Top-1}  $\uparrow$ & \makecell[c]{FID $\downarrow$} & \makecell[c]{MM-Dist $\downarrow$} \\

            \midrule

            \multirow{1}{*}{{\makecell[c]{Token Choice}}}
            
            & \metric{ 0.505}{.006} & \metric{5.095}{.070} & \metric{3.766}{.001} \\


            \multirow{1}{*}{{\makecell[c]{Expert Choice}}}

            & \metric{0.441}{.021} & \metric{8.699}{.140} & \metric{3.796}{.001} \\


            \multirow{1}{*}{{\makecell[c]{Ours}}}

            & \bestmetric{0.512}{.004}  & \bestmetric{4.677}{.069} & \bestmetric{3.762}{.001} \\

            \bottomrule
        \end{tabular}
    }

    \caption{
    Ablation results of the MoE type.
    }
    \label{tab:cmp_moe_routing}
\end{table}

\begin{table}[t]\small
    \centering
    \setlength{\tabcolsep}{0.9mm}{
        \begin{tabular}{l c c c c}
            \toprule
            {\makecell[l]{Expert\\Number}} & \makecell[c]{Total\\Params} & \makecell[c]{R-Precision\\Top-1}  $\uparrow$ & \makecell[c]{FID $\downarrow$} & \makecell[c]{MM-Dist $\downarrow$} \\

            \midrule
            \makecell[l]{None\\(InterGen)} & 182M & \metric{0.371}{.010} & \metric{5.918}{.079} & \metric{5.108}{.014} \\

            \midrule
            
            4 &  164M & \metric{0.494}{.006} & \metric{5.114}{.074} & \metric{3.773}{.001} \\

            8  & 240M & \bestmetric{0.512}{.004}  & \bestmetric{4.677}{.069} & \bestmetric{3.762}{.001} \\

            16 & 391M & \metric{0.507}{.005} & \metric{4.970}{.090} & \metric{3.767}{.003}  \\

            \bottomrule
        \end{tabular}
    }
    \caption{
    Ablation results of the expert number.
    }
    \label{tab:cmp_moe_exp_num}
    
\end{table}

\begin{table}[t]\small
    \centering
    \setlength{\tabcolsep}{0.9mm}{
        \begin{tabular}{l c c c}
            \toprule
            {\makecell[l]{$C^\text{exp}$}} & \makecell[c]{R-Precision Top-1}  $\uparrow$ & \makecell[c]{FID $\downarrow$} & \makecell[c]{MM-Dist $\downarrow$} \\

            \midrule
            
            0.8 & \metric{0.510}{.006} & \metric{4.933}{.075} & \metric{3.766}{.001} \\

            1 & \metric{0.512}{.004}  & \bestmetric{4.677}{.069} & \bestmetric{3.762}{.001} \\

            2 & \bestmetric{0.517}{.007} & \metric{4.878}{.073} & \metric{3.765}{.001} \\

            \bottomrule
        \end{tabular}
    }
    \caption{
    Ablation results of the expectation of the number of experts allocated per feature $C^\text{exp}$.
    }
    \label{tab:cmp_moe_expk}
    
\end{table}

\begin{table}[h]\small
    \centering
    \setlength{\tabcolsep}{0.1mm}{
        \begin{tabular}{l c c c c}
            \toprule
            {Methods} & {\makecell[c]{R-Precision\\Top 1}   $\uparrow$} & {FID $\downarrow$} & {MM-Dist $\downarrow$}  \\

            \midrule
            
            Real motion & \metric{0.511}{.003} & \metric{0.002}{.000} & \metric{2.974}{.008}  \\
            
            \midrule
            
            MDM~\shortcite{tevet2023mdm} & \metric{0.320}{.005} & \metric{0.544}{.044} & \metric{5.566}{.027} \\
            
            \quad+ Ours & \bestmetric{0.434}{.006} & \bestmetric{0.483}{.031} & \bestmetric{2.649}{.009}  \\
            
            \midrule
            
            MLD~\shortcite{chen2023mld} & \metric{0.481}{.003} & \metric{0.473}{.013} & \metric{3.196}{.010} \\
            \quad+ Ours & \bestmetric{0.493}{.002} & \bestmetric{0.398}{.010} & \bestmetric{3.138}{.011} \\
            
            \midrule

            SALAD~\shortcite{hong2025salad} & \metric{0.581}{.003} & \metric{0.076}{.002} & \metric{2.649}{.009}  \\
            
            \quad+ Ours & \bestmetric{0.586}{.003} & \bestmetric{0.069}{.002} & \bestmetric{2.632}{.008} \\
            
            \bottomrule
        \end{tabular}
    }
    \caption{
    Quantitative results on the HumanML3D test set,
    demonstrate the generalization of our InterMoE framework. 
    }
    \label{tab:abl_single_small}
\end{table}

\subsubsection{MoE Analysis}

We conduct a detailed analysis of the Mixture of Expert design, investigating the impact of different MoE paradigms, the number of experts, and different hyperparameters on InterHuman test dataset. The results are summarized in Table~\ref{tab:cmp_moe_routing}, \ref{tab:cmp_moe_exp_num}, and ~\ref{tab:cmp_moe_expk}.
For more qualitative results, please refer to the appendix.

We compare our proposed MoE against two other conventional MoE paradigms: Token-Choice (TC) and Expert-Choice (EC). 
As shown in \ref{tab:cmp_moe_routing}, the results indicate that the standard EC paradigm struggles to effectively leverage the contextual information from the diffusion process during assignment and is limited in fully leveraging the capabilities of the experts. The TC paradigm, by assigning a fixed number of experts to each token, provides a stronger baseline. Our proposed method substantially outperforms both standard paradigms. The significant improvement in FID demonstrates the powerful capability of our Synergistic Router and Dynamic Temporal Selection mechanisms in enhancing the quality of the generated interaction.


We further investigate the trade-off of expert numbers as shown in Table~\ref{tab:cmp_moe_exp_num}.
Our MoE paradigm with a small number of experts yields substantial improvements across all metrics compared to the dense baseline (InterGen). Increasing the number of experts to 8 further boosts performance, achieving the best results on all metrics. However, doubling the experts to 16 results in a slight degradation in performance across all metrics. This suggests that a larger number may introduce redundancy or require more extensive training.

We also investigate the impact of the hyperparameter  $K_e^\text{exp}$, mentioned in Section~\ref{subsec:intermoe}. 
Annotating the expectation of the number of experts allocated per feature as $C^\text{exp}$, we use the value of $C^\text{exp}$ to represent the distinction for different settings of $K_e^\text{exp}$ following common practice by setting 
$$
K_e^\text{exp}=\frac{C^\text{exp} \times \text{Sequence Length}}{\text{Expert Number}} \label{eq:c_exp}
$$
as shown in Table~\ref{tab:cmp_moe_expk}, our model achieves the best performance on both FID and MM-Dist metrics when $C^\text{exp}=1$, indicating optimal generation quality and diversity at this setting. Increasing $C^\text{exp}$ to 2 yields a marginal improvement in R-Precision (text-motion alignment) but degrades the more critical FID score. This may be because forcing more experts to process the same feature can lead to redundancy, while also incurring higher computational costs. Conversely, decreasing $C^\text{exp}$ to 0.8 also yields degradation, which suggests that a too sparse allocation provides insufficient modeling capacity to fully capture the complexity of the target interaction.

\subsubsection{Single Human Motion Generation}

To investigate the generalizability of our framework, we integrated our Dynamic Temporal-Selective MoE paradigm into classic diffusion-based models for single-person motion generation while other hyperparameters remain unchanged. The experimental results, summarized in Table~\ref{tab:abl_single_small}, strongly support this hypothesis. Upon incorporating our MoE, all baseline models exhibit a consistent and significant performance boost.

\section{Conclusion}

In this paper, we present InterMoE, a novel framework for generating 3D human interactions. The core of our work is a new dynamic temporal-selective MoE paradigm. 
By integrating the Synergistic Router and Dynamic Temporal Selection mechanism, InterMoE achieves significant improvements in both individual-specific characteristics and semantic fidelity.
Comprehensive experiments demonstrate that InterMoE surpasses existing state-of-the-art models on several key metrics. Moreover, the excellent performance of our MoE paradigm on single-person tasks underscores the generalizability and broad potential of our framework.

\bibliography{aaai2026}

\appendix


\section{Metrics Detail}
\label{appendix:sec:metrics}

\paragraph{Frechet Inception Distance (FID)} Features are extracted from generated motions and real motions. Subsequently, FID is calculated by comparing the feature distribution of the generated motions with that of the real motions. FID serves as a crucial metric extensively utilized to assess the overall quality of the synthesized motions.

\paragraph{R Precision} For each generated motion, a description pool is created consisting of its ground-truth text description and 31 randomly chosen mismatched descriptions from the test set. Next, the Euclidean distances between the motion and text features of each description in the pool are computed and ranked. We then calculate the average accuracy at the top-1, top-2, and top-3 positions. If the ground truth entry appears among the top-k candidates, it is considered a successful retrieval; otherwise, it is deemed a failure.

\paragraph{MM-Dist} MM distance is calculated as the mean Euclidean distance between the motion feature of each generated motion and the text feature of its corresponding description in the test set.

\paragraph{Diversity} Diversity measures the variance of the generated motions. From the entire set of generated motions, two subsets of the same size $S_d$ are randomly sampled. Their respective sets of motion feature vectors $\{\mathbf{v}_1,...,\mathbf{v}_{S_d}\}$ and $\{\mathbf{v}_1',...,\mathbf{v}_{S_{d}'}\}$ are extracted. The diversity of this set of motions is defined as
\begin{equation*}
    \mathrm{Diversity} = \frac{1}{S_d}\sum_{i=1}^{S_d}\parallel \mathbf{v}_i-\mathbf{v}_i' \parallel_2.
\end{equation*}
$S_d=300$ is used in experiments.

\paragraph{Multi Modality} Multi Modality measures how much the generated motions diversify within the same text. Given a set of motions corresponding to a specific text,  two subsets of the same size $S_l$ are randomly sampled. Their respective sets of motion feature vectors $\{\mathbf{v}_{c,1},...$ $,\mathbf{v}_{c, S_l}\}$ and $\{\mathbf{v}_{c,1}',...,\mathbf{v}_{c, S_l}'\}$ are extracted. The Multi Modality of this motion set is formalized as
\begin{equation*}
    \mathrm{MultiModality} = \frac{1}{C \times S_l} \sum_{c=1}^C \sum_{i=1}^{S_l} \left\| \mathbf{v}_{c,i}-\mathbf{v}'_{c,i} \right\|_2.
\end{equation*}
${S_l}=100$ is used in experiments.

\section{Additional Experiment Results}
\label{appendix:sec:exp-res}

\begin{figure*}[t]
\centering
\includegraphics[width=0.99\textwidth]{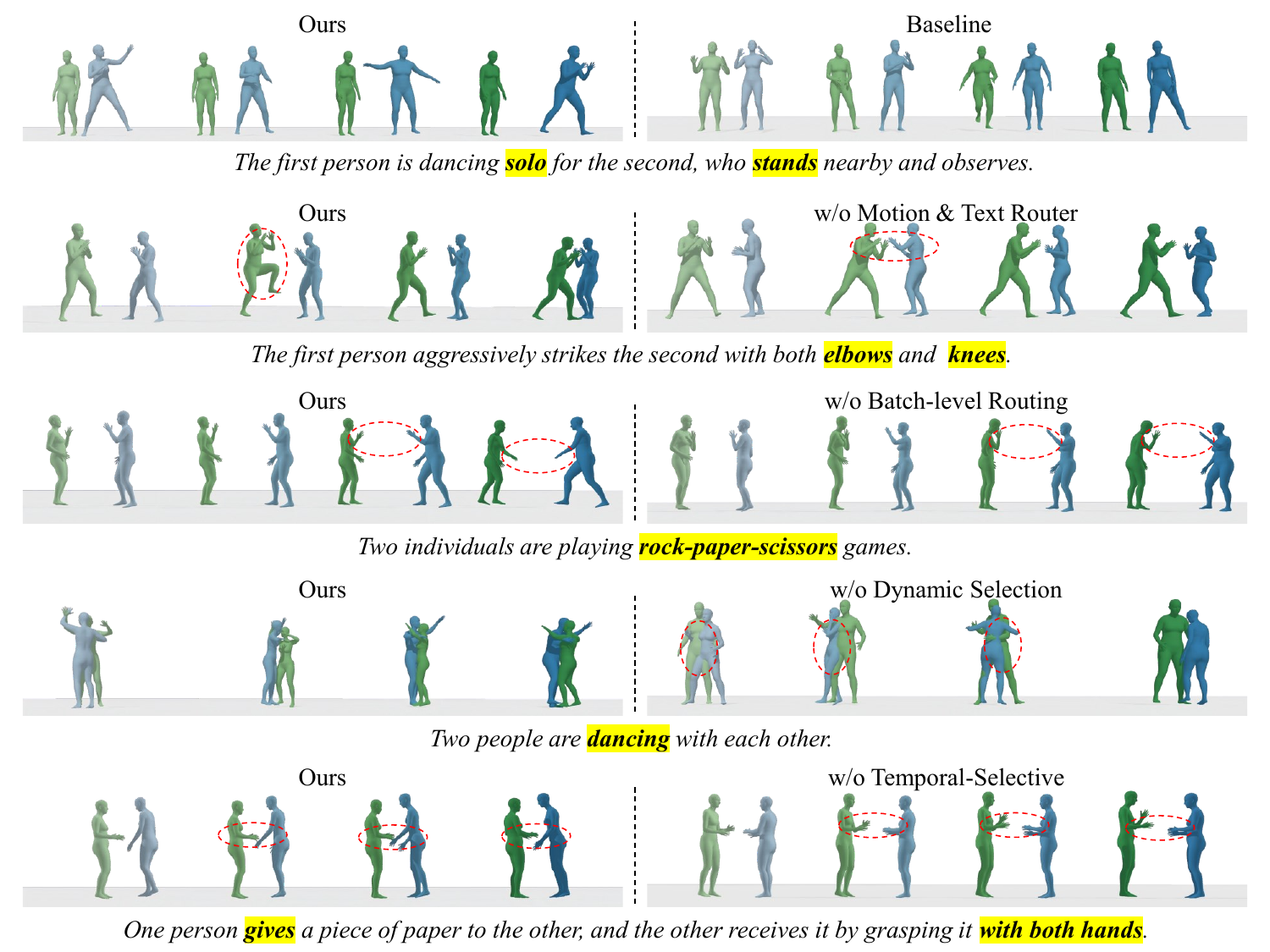}
\caption{
Qualitative results to verify key components of our InterMoE.
}
\label{fig:abl_cmp}
\end{figure*}

\subsection{Ablation study on key components}

We provide additional qualitative results, concerning Table 2 in the main paper, to dissect the contribution of each component within our proposed framework, as shown in Figure~\ref{fig:abl_cmp}.

For the prompt ``The first person is dancing solo for the second, who stands nearby and observes,'', compared to the baseline, our MoE paradigm effectively prevents motion mirroring between individuals, thus preserving individual-specific characteristics. For the prompt ``The first person aggressively strikes the second with both elbows and knees,'', our Motion \& Text Router successfully captures the ``knees'' action described in the text, demonstrating its strong capability for fine-grained semantic understanding. In the ``Two individuals are playing rock-paper-scissors games'' example, removing Batch-Level Routing leads to a general degradation in quality, most notably failing to articulate the intricate hand gestures.
Furthermore, with the aid of Dynamic Selection, ours successfully renders complex hand and foot movements in a ``Two people are dancing with each other" scene while avoiding artifacts like inter-penetration. Finally, for the prompt "One person gives a piece of paper to the other, and the other receives it by grasping it with both hands," our Temporal-Selective mechanism accurately captures the "grasp with both hands" detail, which is missed by the ablated model.

Collectively, these visual comparisons substantiate the efficacy of our design, which demonstrates that through the Dynamic Temporal-Selective MoE, our framework generates high-quality 3D human interactions that exhibit superior semantic fidelity and robust identity preservation.

\begin{figure*}[t]
\centering
\includegraphics[width=0.99\textwidth]{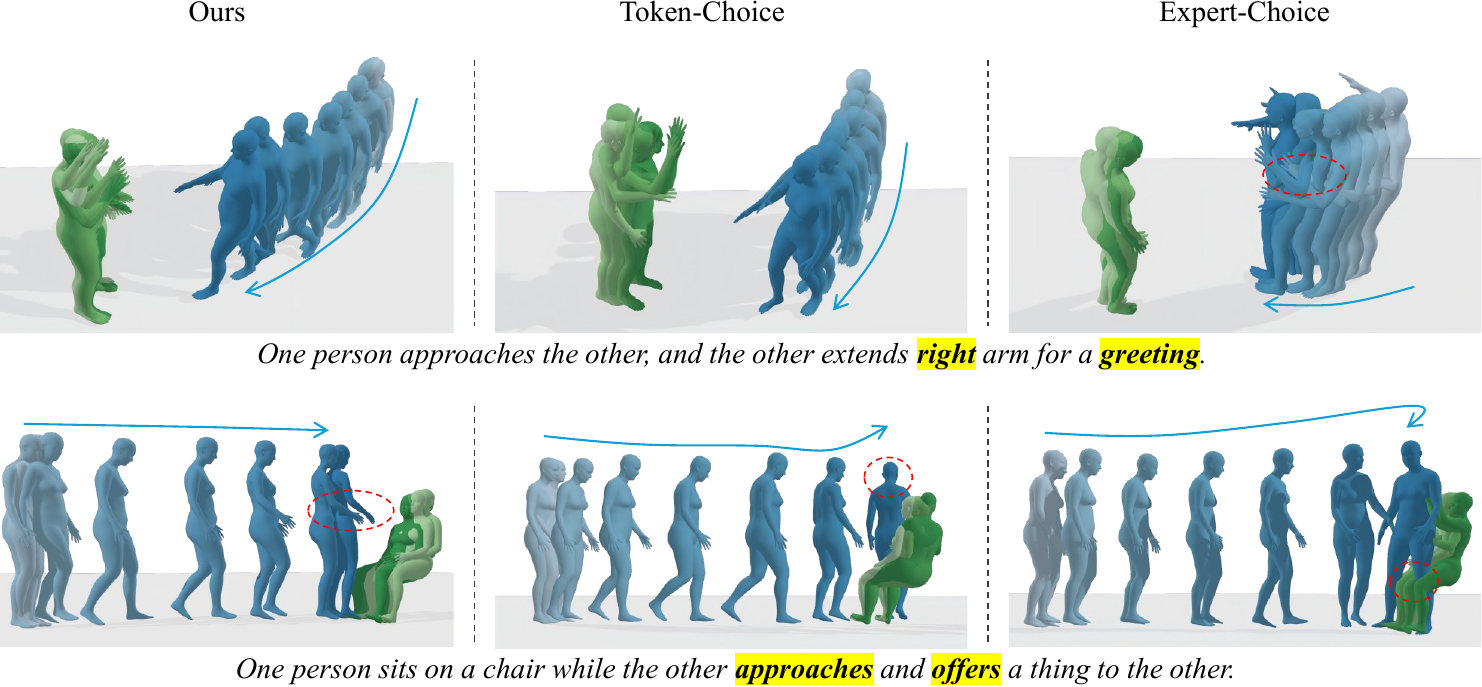}
\caption{
Qualitative comparisons among the different MoE types. Arrowed lines mark the trajectories of motion.
}
\label{fig:abl_moe}
\end{figure*}

\subsection{Ablation study on different MoE types}

We provide further qualitative comparisons of our MoE paradigm against Token-Choice (TC)~\cite{fei2024scaling} and a vanilla Expert-Choice (EC)~\cite{sun2024ecdit} in Figure~\ref{fig:abl_moe}, concerning Table 3 in the main paper.

For the prompt, ``One person approaches the other, and the other extends right arm for a greeting,'' ours accurately synthesizes the approaching relationship and the key gesture of extending the right arm. In contrast, the TC produces illogical character orientations, while the EC fails to correctly articulate the “extends right arm” action. Similarly, in the scenario ``One person sits on a chair while the other approaches and offers a thing to the other,'' ours successfully renders both the “approaching” dynamic and the nuanced “offer” gesture. The TC result, however, exhibits unnatural relative orientations after the approach, and the EC variant suffers from severe inter-penetration artifacts. Furthermore, both competing methods lack the necessary fidelity to express the details of the “offer” action.

These comparisons collectively demonstrate that, unlike conventional MoE implementations, ours generates interaction sequences that are both semantically precise and kinematically plausible.

\section{Causal-Skeletal VAE}

\subsection{Architecture detail}

As mentioned in the main paper, in section 3.1, the skeletal convolution~\cite{hong2025salad} is a graph convolution over the joint dimension, which is defined as follows:
\begin{align*}
\text{Skeletal Conv}({m}^{j}) := \mathbf{\Theta}_{1} ({m}^{j}) + \frac{1}{| \mathcal{N} (j) |} \sum_{n \in \mathcal{N} (j)} \mathbf{\Theta} ({m}^n)
\end{align*}
where $m^{j}$ is the motion feature of the $j$-th joint, $\Theta_{\{1,2\}}$ represents a linear feed-forward layer, and $ \mathcal{N} (j) $ denotes the indices of joints neighboring to joint $j$.

\begin{figure}[ht]
\centering
\includegraphics[width=0.4\columnwidth]{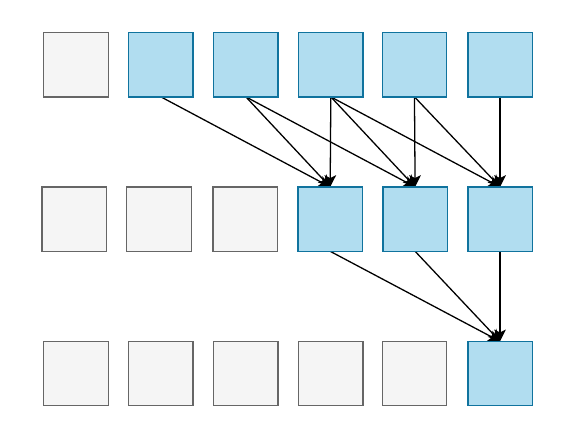}
\caption{Illustration of the causal convolution.}
\label{fig:vae_causal}
\end{figure}

We employ 2D causal convolution~\cite{yu2023language} to effectively model the inherent temporal and causal dependencies in motion data. This approach enables continuous motion compression while explicitly enforcing causal constraints on the sequential representation. Causality is strictly maintained through an asymmetric temporal padding scheme. Specifically, for a convolution layer with kernel size $k_t$, stride $s_t$, and dilation rate $d_t$, we pad $(k_t - 1) \times d_t + (1 - s_t)$ frames to the beginning of the sequence. As shown in~\ref{fig:vae_causal}, this ensures that the output at any given timestep is influenced only by past and current, allowing the model to more effectively learn the causal structure within the motion data.

\begin{figure}[h]
\centering
\includegraphics[width=0.99\columnwidth]{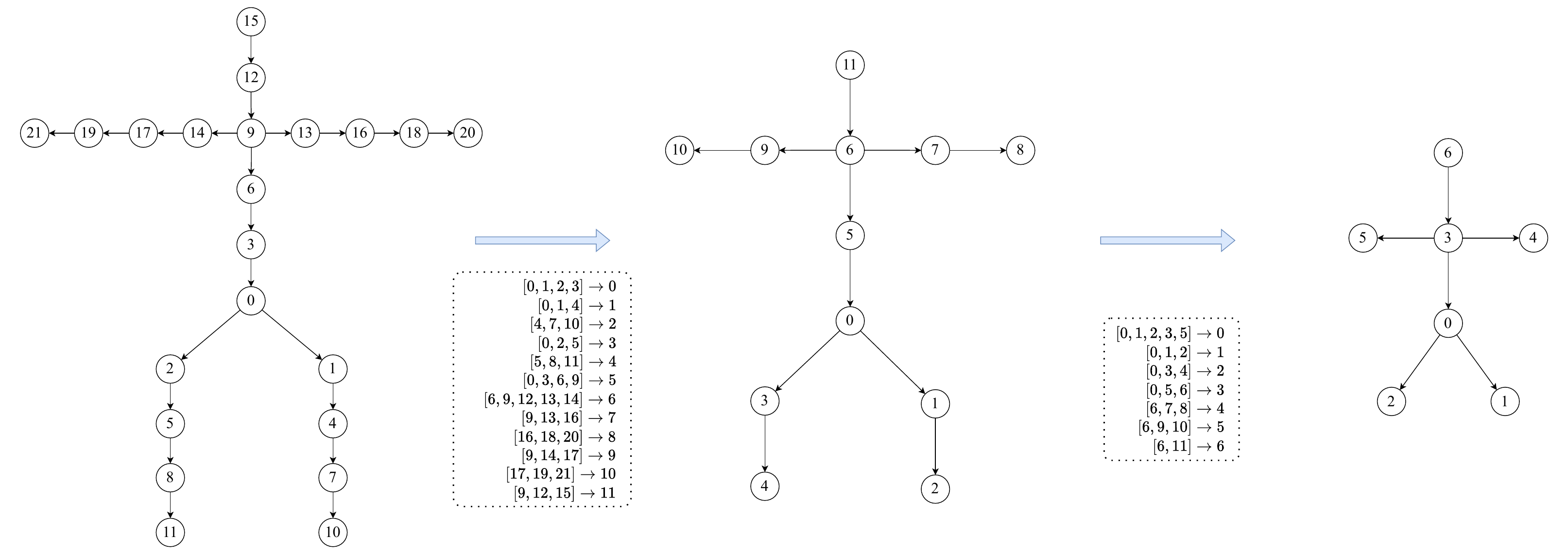}
\caption*{(a) InterHuman}
\includegraphics[width=0.99\columnwidth]{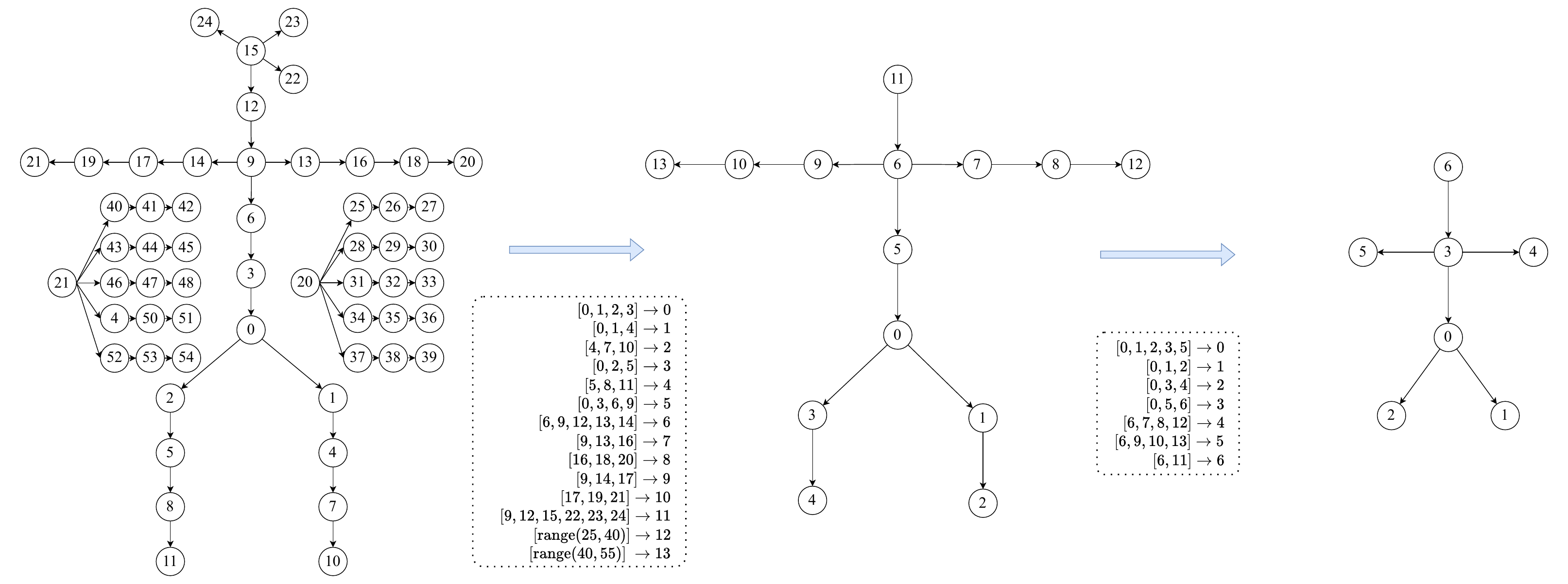}
\caption*{(b) Inter-X}

\caption{Illustration of the skeletal pooling process for the InterHuman and Inter-X datasets. The original skeleton (left) is progressively abstracted by pooling adjacent joints (middle and right). The notation $ [ \ ] \rightarrow i $ indicates the abstracted joint index and the set of original joints that are grouped together. The unpooling layers operate in the reverse order to restore the skeletal resolution.}
\label{fig:vae_pooling}
\end{figure}

For the pooling layers, we apply the average pooling across both Skeletal and temporal dimensions. Specifically, temporal features are pooled with a kernel of size 2 and a stride of 2, while skeletal pooling reduces the number of joints by summarizing adjacent joints. In contrast, the unpooling layers perform the inverse operation of the pooling. 
To increase the skeletal resolution, we recreate unpooled joints by summing the features from the corresponding pooled joints. The temporal resolution is improved by upsampling the features along the temporal dimension using linear interpolation. The downsampling and upsampling for each dataset are visualized in Figure~\ref{fig:vae_pooling}.

\subsection{Loss terms for the training stage}
The VAE is trained with the following objectives: 
\begin{align}
\mathcal{L}_\text{VAE} = \mathcal{L}_m + 
\lambda_\text{pos} \mathcal{L}_\text{pos} + 
\lambda_\text{vel} \mathcal{L}_\text{vel} + 
\lambda_\text{kl}  \mathcal{L}_\text{kl},
\end{align}
where $\mathcal{L}_m $, $\mathcal{L}_\text{pos}$, $\mathcal{L}_\text{vel}$ are $L1$ reconstruction loss terms of the motion features, joint positions, and joint velocities, while $\mathcal{L}_\text{kl}$ is the Kullback–Leibler (KL) divergence regularization that encourages a structured latent space. 
In experiments, we set  $\lambda_\text{pos}=0.5$, $\lambda_\text{vel}=0.5$, $\lambda_\text{kl}=0.02$.

\section{Dynamic Selection Analysis}

\begin{figure}[ht]
\centering
\includegraphics[width=0.99\columnwidth]{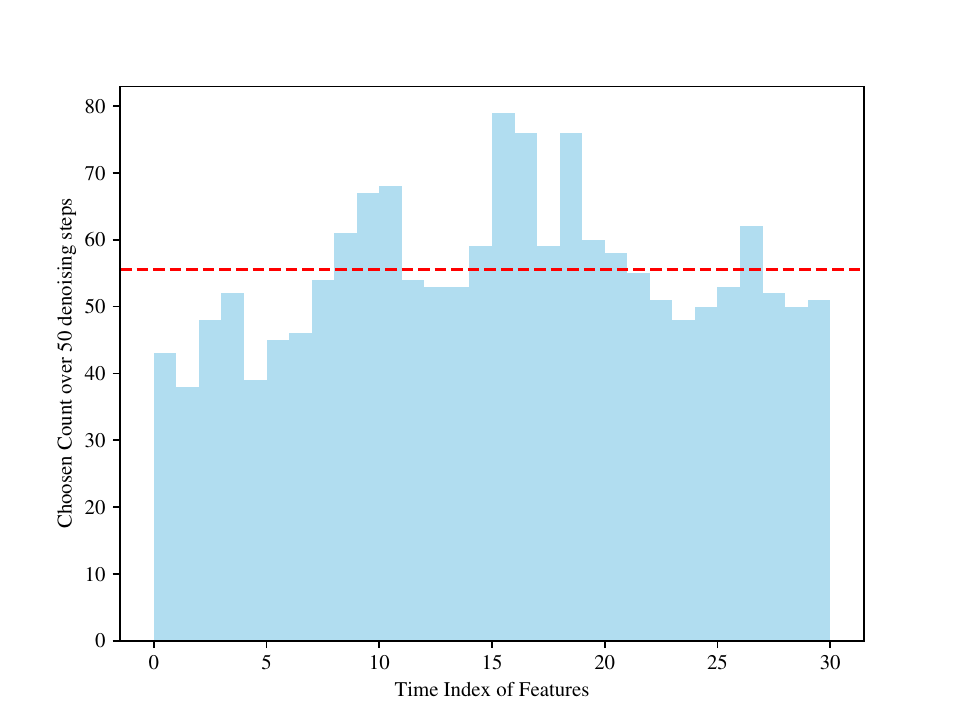}
\caption{Counting of selected time of the temporal feature during inference (50 denoising steps). We infer a 120s interaction ``Two men dancing together.''. Note that after being encoded by the VAE, the length of the temporal dimension is 30.}
\label{fig:feature_count}
\end{figure}

To validate the dynamic selection behavior of our MoE mechanism, we visualize the expert selection frequency for each temporal motion feature during the inference process, as depicted in Figure~\ref{fig:feature_count}. For this analysis, we set the number of experts allocated per feature $C^\text{exp}$ to 1 ($C^\text{exp}$ as mentioned in the main paper section 4.3).

We observe that across a 50-step denoising process, the total selection count per feature averages approximately 50. This result aligns perfectly with our theoretical target ($C^\text{exp} = 1$), confirming that our mechanism is well-calibrated and functions as designed. More importantly, the visualization reveals a distinct, non-uniform selection pattern, indicating that the experts are indeed making discriminative choices about which temporal features to process. This ability to dynamically prioritize and focus on the most salient features is fundamental to the effectiveness of our method, demonstrating the superiority of its dynamic selection strategy.

\section{User Study}
\begin{table}[h]\small
    \centering
    \setlength{\tabcolsep}{0.5mm}{
        \begin{tabular}{l c c c }
            \toprule
            \makecell[c]{Methods} & \makecell[c]{Individual-Specific\\Characteristic} & \makecell[c]{Semantic \\ Fidelity} & \makecell[c]{Overall \\ Quality} \\

            \midrule

            InterMask~\shortcite{javedintermask} & 3.54 & 3.36 & 3.07 \\ 
            TIMotion~\shortcite{wang2025TIMotion} & 2.32 & 3.44 & 2.98 \\ 
            Ours & 4.16 & 4.45 & 4.20 \\ 

            \bottomrule
        \end{tabular}
    }

    \caption{
    User study results.
    }
    \label{tab:user_study}
\end{table}
To conduct a more comprehensive and objective evaluation of our framework, we performed a user study. In this study, we compared our InterMoE against two leading methods: InterMask~\cite{javedintermask} and TIMotion~\cite{wang2025TIMotion}. We generated 8 distinct interaction sequences for each method from a shared set of text prompts. We then gathered responses from 30 participants, who were asked to rate each sequence based on three criteria from 0 to 5: (1) Individual-Specific Characteristics; (2) Semantic Fidelity; and (3) Overall Quality. The results, summarized in Table~\ref{tab:user_study}, indicate a strong user preference for our InterMoE across all criteria. The findings particularly highlight the superiority of our InterMoE in preserving individual-specific characteristics and maintaining high fidelity.

\section{Limitations and Future Work}
While InterMoE achieves strong results, several areas remain for future exploration. 
First, in line with other data-driven approaches, our framework does not explicitly enforce physical constraints, which can result in occasional, minor artifacts such as inter-penetration, jitter, or foot-sliding. 
Second, our work focuses on human-human interaction; extending the framework to model human-object and human-animal interactions presents a valuable direction for future research.
Finally, efficiently extending the current framework to support long-horizon generation and complex, multi-agent (\emph{i.e}., more than two) interactions are important direction for future exploration.

\end{document}